\documentclass{article}

\usepackage{arxiv}

\usepackage[utf8]{inputenc} 
\usepackage[T1]{fontenc}    
\usepackage{hyperref}       
\usepackage{url}            
\usepackage{booktabs}       
\usepackage{amsfonts}       
\usepackage{nicefrac}       
\usepackage{microtype}      
\usepackage{lipsum}

\usepackage{hyperref}
\usepackage{cleveref}

\usepackage{graphicx}
 \usepackage{algorithm}
\usepackage{algorithmic}

\usepackage{calrsfs}
\DeclareMathAlphabet{\pazocal}{OMS}{zplm}{m}{n}

\newcommand{\Ea}{\pazocal{E}}
\newcommand{\Fa}{\pazocal{F}}
\newcommand{\Aa}{\pazocal{A}}
\newcommand{\Ra}{\pazocal{R}}
\newcommand{\Sa}{\pazocal{S}}

\newcommand{\Ca}{\pazocal{C}}

\usepackage{amssymb,amsmath}

\title{An Imprecise Probability Approach for Abstract Argumentation based on Credal Sets}

\author{
  Mariela Morveli-Espinoza \\
  Graduate Program in Electrical and Computer Engineering (CPGEI),\\
 Federal University of Technology - Paran\'{a} (UTFPR),
 Curitiba - Brazil\\
  \texttt{morveli.espinoza@gmail.com} \\
   \And
   Juan Carlos Nieves \\
Department of Computing Science of Ume{\aa}  University, \\
 Ume{\aa}  - Sweden
 \texttt{jcnieves@cs.umu.se} \\
   \And
   Cesar Augusto Tacla \\
  Graduate Program in Electrical and Computer Engineering (CPGEI),\\
 Federal University of Technology - Paran\'{a} (UTFPR),
 Curitiba - Brazil\\
   \texttt{tacla@utfpr.edu.br} \\
}

\newtheorem{definition}{Definition}
\newtheorem{proposition}{Proposition}
\newtheorem{example}{Example}
\newtheorem{proof}{Proof}

\begin{document}

\maketitle

\begin{abstract}
Some abstract argumentation approaches consider that arguments have a degree of uncertainty, which impacts on the degree of uncertainty of the extensions obtained from a abstract argumentation framework (AAF) under a semantics. In these approaches, both the uncertainty of the arguments and of the extensions are modeled by means of precise probability values. However, in many real life situations the exact probabilities values are unknown and sometimes there is a need for aggregating the probability values of different sources. In this paper, we tackle the problem of calculating the degree of uncertainty of the extensions considering that the probability values of the arguments are imprecise. We use credal sets to model the uncertainty values of arguments and from these credal sets, we calculate the lower and upper bounds of the extensions. We study some properties of the suggested approach and illustrate it with an scenario of decision making.

\end{abstract}


\section{Introduction}
The AAF that was introduced in the seminal paper of Dung \cite{dung1995acceptability} is one of the most significant developments in the computational modelling of argumentation in recent years. The AAF is composed of a set of arguments and a binary relation encoding attacks between arguments. Some recent approaches on abstract argumentation assign uncertainty to the elements of the AAF to represent the degree of believe on arguments or attacks. Some of these works assign uncertainty to the arguments (e.g., \cite{dung2010towards}\cite{hunter2012some}\cite{thimm2012probabilistic}\cite{hunter2013probabilistic}\cite{hunter2014probabilistic}\cite{gabbay2015probabilistic}\cite{riveret2015probabilistic}\cite{thimm2017probabilities}), others to the attacks (e.g., \cite{hunter2014probabilistic}), and others to both arguments and attacks (e.g., \cite{li2011probabilistic}). These works use precise probability approaches to model the uncertainty values. However, precise probability approaches have some limitations to quantify epistemic uncertainty, for example, to represent group disagreeing opinions. These can be better represented by means of imprecise probabilities, which use lower and upper bounds instead of exact values to model the uncertainty values. 

\begin{figure}[!htb]
	\centering
		\includegraphics[width=0.6\textwidth]{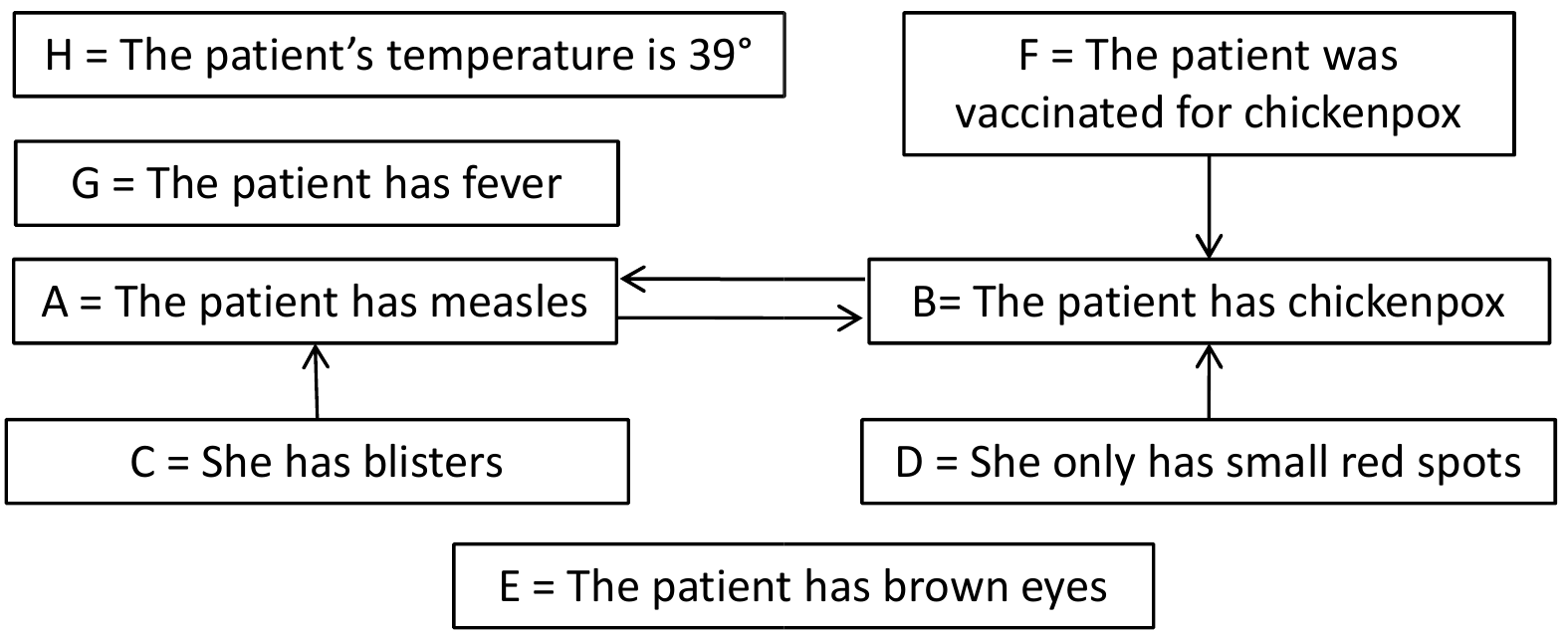} 
	\caption{Argumentation graph for the discussion about the diagnose of a patient.}
	\label{conversa}
\end{figure}

For a better illustration of the problem, consider a discussion between a group of medicine students (agents). The discussion is about the diagnose of a patient. In this context, arguments represent the student's opinions and the attacks represent the disagreements between such opinions. Figure \ref{conversa} shows the argumentation graph where nodes represent arguments and edges the attacks between arguments. In the graph, two arguments represent two possible diagnoses namely measles and chickenpox, there is an argument against measles and two arguments against chickenpox, and there are three arguments that have no attack relations with the rest of arguments.

Suppose that each opinion -- i.e., argument -- has a probability value between 0 and 1 that represents the degree of believe of each student. Since there is more than one opinion, this means that each argument has associated a set of probability values.
Thus, we cannot model these degrees of believe by means of an unique probability value (precise probability value), what we need is to represent a range of the possible degrees of believe.  

To the best of our knowledge, there is no work that models the uncertainty values of arguments by using an imprecise probability approach. Therefore, we aim to propose an approach for abstract argumentation in which the uncertainty of the arguments is modeled by an imprecise probability value. Thus, the research questions that are addressed in this paper are: 

\begin{enumerate}
\item How to model the imprecise uncertainty values of arguments?
\item In abstract argumentation, several semantics have been proposed, which return sets of arguments -- called extensions -- whose basic characteristic is that these arguments do not attack each other, i.e. they are consistent. The fact that the arguments that belong to an extension are uncertain, causes that such extension also has a degree of uncertainty. How to calculate the lower and upper bounds of extensions?

\end{enumerate}

In addressing the first question, we use credal sets to model the uncertainty values of arguments. Regarding the second question, we base on the credal sets of the arguments to calculate the uncertainty values of extensions obtained under a given semantics. These values are represented by lower and upper bounds. The way to aggregate the credal sets depends on a causal relation between the arguments. 

 Next section gives a brief overview on credal sets and abstract argumentation. In Section \ref{AAF-credal}, we present the AAF based on credal sets and the causality graph concept, which are the base for the calculation of the upper and lower bounds of extension. This calculation is tackled in Section \ref{LU-ext}. We study the main properties of our approach in Section \ref{properties}. Related work is presented in Section \ref{relacionado}. Finally, Section \ref{conclus} is devoted to conclusions and future work.

\section{Background}
\label{teoria}

In this section, we revise concepts of credal sets and abstract argumentation.

\subsection{Credal sets}

Assume that we have a finite set of events $\mathbb{E}= \{E_1, ..., E_n\}$ and a probability distribution $p$ on this set, where $p$ is a mapping $p: \mathbb{E} \rightarrow [0,1]$. According to Levi \cite{levi1983enterprise}, a closed convex set of probability distributions $p$ is called a credal set. Given an event $E$, a credal set for $E$ -- denoted $K(E)$ -- is a set of probability distributions about this event and $\mathbb{K}=\{K(E_1),...,K(E_n)\}$ denotes a set of all credal sets. Every credal set has the same number of elements. In this work, we assume that the cardinality of the credal sets of $\mathbb{K}$ is the same (let us denote it by $m$); moreover, we assume that $p_i(E)$ denotes the suggested probability of the agent $i$ w.r.t the event $E$ such that $1 \leq i \leq m$ and $E \in \mathbb{E}$. Given a credal set $K(E)$, the lower and upper bounds for event $E$ are determined as follows:

\textbf{Lower probability:} $\underline{P}(E)=inf \{p(E): p(E) \in K(E)\}$ \hspace*{1.2cm} (1)\\
\indent\textbf{Upper probability:} $\overline{P}(E)=sup \{p(E): p(E) \in K(E)\}$

Given $l$ events $\{E_1, ..., E_l\} \subseteq \mathbb{E}$ and their respective credal sets $K(E_1)=\{p_1(E_1),..., p_m(E_1)\}, ..., K(E_l)=\{p_1(E_l),..., p_m(E_l)\}$. If $\{E_1, ..., E_l\}$ are independent events, the lower and upper probabilities are defined as follows:

\hspace*{0.53cm}$\underline{P}(\{E_1, ..., E_l\})= min_{1 \leq j \leq m} \{ \prod_{i=1 }^{i \leq l} p_j(E_i)\}$ where $p_j \in K(E_i)$  \hspace*{0.4cm} (2)\\
\hspace*{0.53cm}$\overline{P}(\{E_1, ..., E_l\})= max_{1 \leq j \leq m} \{ \prod_{i=1 }^{i \leq l} p_j(E_i)\}$

On the other hand, when the independence relation is not assumed, the first step is to calculate a credal set for $\{E_1, ..., E_l\}$ as follows:

$K(\{E_1, ..., E_l\})= \{p_E | p_E = min_{1 \leq j \leq m} \{p_j(E_1), ... p_j(E_l)\}\}$ where $p_j(E_i) \in K(E_i)$ \hspace*{2cm}(3)

Based on $K(\{E_1, ..., E_l\}) $, we obtain the lower and upper probabilities:

\indent $\underline{P}(\{E_1, ..., E_l\})= min (K(\{E_1, ..., E_l\})) $  \hspace*{4cm}(4)\\
\indent$\overline{P}(\{E_1, ..., E_l\})= max (K(\{E_1, ..., E_l\}))$

\begin{example} \label{exm-1}Let $\{E_1, E_2,E_3\}$ be three events and $K(E_1)=\{p_1(E_1), p_2(E_1), p_3(E_1)\}, K(E_2)=\{p_1(E_2), p_2(E_2),\break p_3(E_2)\},$ and $K(E_3)=\{p_1(E_3), p_2(E_3), p_3(E_3)\}$ their respective credal sets. Next table shows the values of the probability distributions for each event.

\begin{center}
\begin{tabular}{| c | c | c | c |}
\hline 
 & $E_1$ & $E_2$ & $E_3$ \\ 
\hline 
$p_1$ & 0.3 & 0.5 & 0.75 \\ 
\hline 
$p_2$ & 0.6 & 0.7 & 0.55 \\ 
\hline 
$p_3$ & 0.45 & 0.65 & 0.8 \\ 
\hline 
\end{tabular} 
\end{center}

Assuming that $E_1, E_2,$ and $E_3$ are independent, the lower and upper probabilities of $(E_1,E_2,E_3)$ are calculated as follows: $\underline{P}(E_1,E_2,E_3)= min \{0.3 \times 0.5 \times 0.75, 0.6 \times 0.7 \times 0.55, 0.45 \times 0.65 \times 0.8\} = min \{0.1125, 0.231, 0.234\}$; hence $\underline{P}(E_1,E_2,E_3)= 0.1125$ and $\overline{P}(E_1,E_2,E_3)= max \{0.1125, 0.231, 0.234\}= 0.234$. 

On the other hand, if we assume that $E_1, E_2,$ and $E_3$ are not independent, then the lower and upper probabilities are calculated as follows: $K(E_1, E_2, E_3)=\{min\{0.3,0.5,0.75\}, min \{0.6,0.7,0.5\}, min\{0.45,0.65,0.8\} \}=\{0.3, 0.55,0.45\}$. Thus, $\underline{P}(E_1,E_2,E_3)=0.3$ and $\overline{P}(E_1,E_2,E_3)=0.55$.

\end{example}

\subsection{Abstract argumentation}

In this subsection, we will recall basic concepts related to the AAF defined by Dung \cite{dung1995acceptability}, including the notion of acceptability and the main semantics.

\begin{definition}\label{def-dung} \textbf{(Abstract AF)} An abstract argumentation framework $\Aa\Fa$ is a tuple $\Aa\Fa = \langle \mathtt{ARG}, \Ra \rangle$ where $\mathtt{ARG}$ is a finite set of arguments and $\Ra$ is a binary relation $\Ra$ $\subseteq \mathtt{ARG} \times \mathtt{ARG}$ that represents the attack between two arguments of $\mathtt{ARG}$, so that $(A,B) \in \Ra$ denotes that the argument $A$ attacks the argument $B$.

\end{definition}

Next, we introduce the concepts of conflict-freeness, defense, admissibility and the four semantics proposed by Dung \cite{dung1995acceptability}.

\begin{definition}\label{def-semantics} \textbf{(Argumentation Semantics)}\label{basicosarg} Given an argumentation framework $\Aa\Fa=\langle \mathtt{ARG}, \Ra \rangle$ and a set $\Ea \subseteq \mathtt{ARG}$:
\begin{itemize}
\item $\Ea$ is \textit{conflict-free} if $\forall A, B \in \Ea$, $(A,B) \not\in \Ra$.
\item $\Ea$ \textit{defends} an argument $A$ iff for each argument $B \in \mathtt{ARG}$, if $(B, A) \in \Ra$, then there exist an argument $C \in \Ea$ such that $(C,B) \in \Ra$.
\item $\Ea$ is \textit{admissible} iff it is conflict-free and defends all its elements. 
\item A conflict-free $\Ea$ is a \textit{complete extension} iff we have $\Ea = \{A | \Ea$ defends $A\}$.
\item $\Ea$ is a \textit{preferred extension} iff it is a maximal (w.r.t set inclusion) complete extension.
\item $\Ea$ is a \textit{grounded extension} iff it is the smallest (w.r.t set inclusion) complete extension.
\item $\Ea$ is a \textit{stable extension} iff $\Ea$ is conflict-free and $\forall A \in \mathtt{ARG}$ and $A \not\in \Ea$, $\exists B \in \Ea$ such that $(B,A) \in \Ra$.
\end{itemize}

\end{definition}

In this article, there is a set of agents that give their opinions (degrees of belief) regarding each argument in $\mathtt{ARG}$ by means of probability distributions. The set of arguments can be compared with the events of set $\mathbb{E}$; hence, we can say that $\mathbb{E} = \mathtt{ARG}$. The number of agents that give their opinions determines the cardinality of credal sets. Thus, given $m$ agents and an argument $A \in \mathtt{ARG}$, the credal set for $A$ is represented by $K(A)=\{p_1(A), ..., p_m(A)\}$. Finally, $\mathbb{K}$ denotes all the credal sets of the arguments in $\mathtt{ARG}$.

\section{The Building Blocks}
\label{AAF-credal}

In this section, we present the definitions of AAF based on credal sets and causality graph. These concepts are important for the calculation of the lower and upper bounds of extensions.

We use credal sets to model the opinions (degrees of belief) of a set of agents about a set of arguments. Thus, each argument in an AAF has associated a credal set, which contains probability distributions that represent the opinions of the agents about it.

\begin{definition} \textbf{(Credal Abstract Argumentation Framework)} An  AAF based on credal sets is a tuple $\Aa\Fa_{\Ca\Sa}=\langle \mathtt{ARG}, \Ra, \mathbb{K},f_{\Ca\Sa}\rangle$ where (i) $\mathtt{ARG}$ is a set of arguments, (ii) $\Ra$ is the attack relation presented in Definition \ref{def-dung}, (iii) $\mathbb{K}$ is a set of credal sets, and (iv) $f_{\Ca\Sa}: \mathtt{ARG} \rightarrow \mathbb{K}$ maps a credal set for each argument in $\mathtt{ARG}$. 

\end{definition}

 Recall that the cardinality of every credal set depends on the number of agents. Since all the agents give their opinions about all the arguments, all the credal sets have the same number of elements. 

\begin{definition} \textbf{(Agent's opinions)} Let $\Aa\Fa_{\Ca\Sa}=\langle \mathtt{ARG}, \Ra, \mathbb{K},f_{\Ca\Sa}\rangle$ be a Credal AAF and $\mathbb{AGT}=\{ag_1, ..., ag_m\}$  a set of agents. The opinion $p_i$ of an agent $ag_i$ (for $1 \leq i \leq m$) is ruled as follows:

1.  If $A \in \mathtt{ARG}$, there is $p_i(A) \in K(A)$ where $K(A)  \in \mathbb{K}$.\\
\indent2. $\forall A \in \mathtt{ARG}$, $0 \leq p_i(A) \leq 1$.

\end{definition}

Regarding the probability values given to the arguments, it is important to consider the notion of rational probability distribution given in \cite{hunter2013probabilistic}. According to Hunter \cite{hunter2013probabilistic}, if the degree of belief in an argument is high, then the degree of belief in the arguments it attacks is low. Thus, a probability function $p$ is rational for an $\Aa\Fa_{\Ca\Sa}$ iff for each $(A,B) \in \Ra$, if $p(A) > 0.5$ then $p(B)\leq 0.5$ where $p(A) \in K(A)$ and $p(B) \in K(B)$.

\begin{example} \label{ejm-2} Consider that $\mathbb{AGT}=\{ag_1, ag_2, ag_3, ag_4\}$. The Credal AAF for the example given in Introduction is $\Aa\Fa_{\Ca\Sa}=\langle \mathtt{ARG}, \Ra, \mathbb{K},f_{\Ca\Sa}\rangle$ where:

- $\mathtt{ARG}= \{A,B, C,D,E,F,G.H\}$\\
\indent- $\Ra=\{(A,B), (B,A), (F,B), (D,B), (C,A)\}$ \\
\indent- $\mathbb{K}=\{K(A),K(B), K(C), K(D), K(E), K(F), K(G), K(H)\}$. The table below shows the credal set of each argument\\
\indent- $f_{\Ca\Sa}(A)=K(A)$, $f_{\Ca\Sa}(B)=K(B)$, ..., $f_{\Ca\Sa}(H)=K(H)$  

\begin{center}
\begin{tabular}{|c|c|c|c|c|c|c|c|c|}
\hline 
 & $K(A)$ & $K(B)$ & $K(C)$ & $K(D)$ & $K(E)$ & $K(F)$ & $K(G)$ & $K(H)$ \\ 
\hline 
$p_1$ & 0.2 & 0.8 & 0.2 & 0.75 & 0.8 & 0.75 & 0.7 & 0.8 \\ 
\hline 
$p_2$ & 0.7 & 0.25 & 0.75  & 0.15 & 0.65 & 0.2 & 0.8 & 0.9\\ 
\hline 
$p_3$ & 0.55 & 0.45 & 0.4 & 0.5 & 0.8 & 0.55 & 1 & 1 \\ 
\hline 
$p_4$ & 0.75 & 0.1 & 0.2 & 0.8 & 0.7 & 0.8 & 0.9 & 0.9 \\ 
\hline 
\end{tabular} 
\end{center}

\end{example}

In a Credal AAF, besides the attack relation between the arguments, there may be a causality relation between them. To make this discussion more concrete, consider the following conflict-free sets:

\begin{itemize}
\item $\{G,E\}$: Having fever does not have to do with the eyes' color of the patient and vice-verse, so there is no relation between these arguments. This means that they are independent from each other. 

\item $\{A,G\}$ and $\{A,F\}$: In both cases the arguments are related in some way. In the first case, having fever ($G$) is a symptom of (causes) measles ($A$) and in the second case, the fact that the patient is vaccinated for chickenpox ($F$) causes that he may have measles and not chickenpox ($A$). 

\end{itemize}

\begin{definition} \textbf{(Causality Graph)} Let $\Aa\Fa_{\Ca\Sa} = \langle \mathtt{ARG}, \Ra, \mathbb{K},f_{\Ca\Sa} \rangle$ be a Credal AAF, a causality graph $\mathbb{C}$ is a tuple $\mathbb{C} = \langle \mathtt{ARG}, \Ra_{\mathtt{CAU}} \rangle$ such that:

\begin{itemize}
\item[i] $\mathtt{ARG}=\mathtt{ARG_{\leftarrow}} \cup \mathtt{ARG_{\rightarrow}} \cup \mathtt{ARG_{\circ}}$ is a set of arguments,
\item[ii] $\Ra_{\mathtt{CAU}} \subseteq \mathtt{ARG} \times \mathtt{ARG}$ represents a causal relation between two arguments of $\mathtt{ARG}$ (the existence of this relation depends on the domain knowledge), such that $(A,B) \in \Ra_{\mathtt{CAU}}$ denotes that argument $A$ causes argument $B$. It holds that if $(A,B) \in \Ra$, then $(A,B) \notin \Ra_{\mathtt{CAU}}$ and $(B,A) \notin \Ra_{\mathtt{CAU}}$,
\item[iii] $\mathtt{ARG_{\leftarrow}}=\{B | (A,B) \in \Ra_{\mathtt{CAU}}\}$, $\mathtt{ARG_{\rightarrow}}= \{A|(A,B) \in \Ra_{\mathtt{CAU}}\}$, and $\mathtt{ARG_{\circ}}=\{C| C \in \mathtt{ARG} - (\mathtt{ARG_{\leftarrow}} \cup \mathtt{ARG_{\rightarrow}})\}$,
\item[iv] $\mathtt{ARG_{\leftarrow}}$ and $\mathtt{ARG_{\rightarrow}}$ are not necessarily pairwise disjoint; however, $(\mathtt{ARG_{\leftarrow}} \cup \mathtt{ARG_{\rightarrow}})$ $ \cap \mathtt{ARG_{\circ}}=\emptyset$.

\end{itemize}
\end{definition}

 \begin{example} \label{ejm-causa}A causality graph for the Credal AAF of Example \ref{ejm-2} is $\mathbb{C} = \langle \{A,B,C, D, E,F, G,H\},\break \{(D,A), (F,A), (H,A), (G,A), (H,G), (G,B), (C,B)\} \rangle$ (see Figure \ref{dep}), where $\mathtt{ARG_{\leftarrow}}=\{A,B,G\}$, $\mathtt{ARG_{\rightarrow}}=\{D,F,H, G,C\}$, and $\mathtt{ARG_{\circ}}=\{E\}$.

\begin{figure}[!htb]
	\centering
		\includegraphics[width=0.45\textwidth]{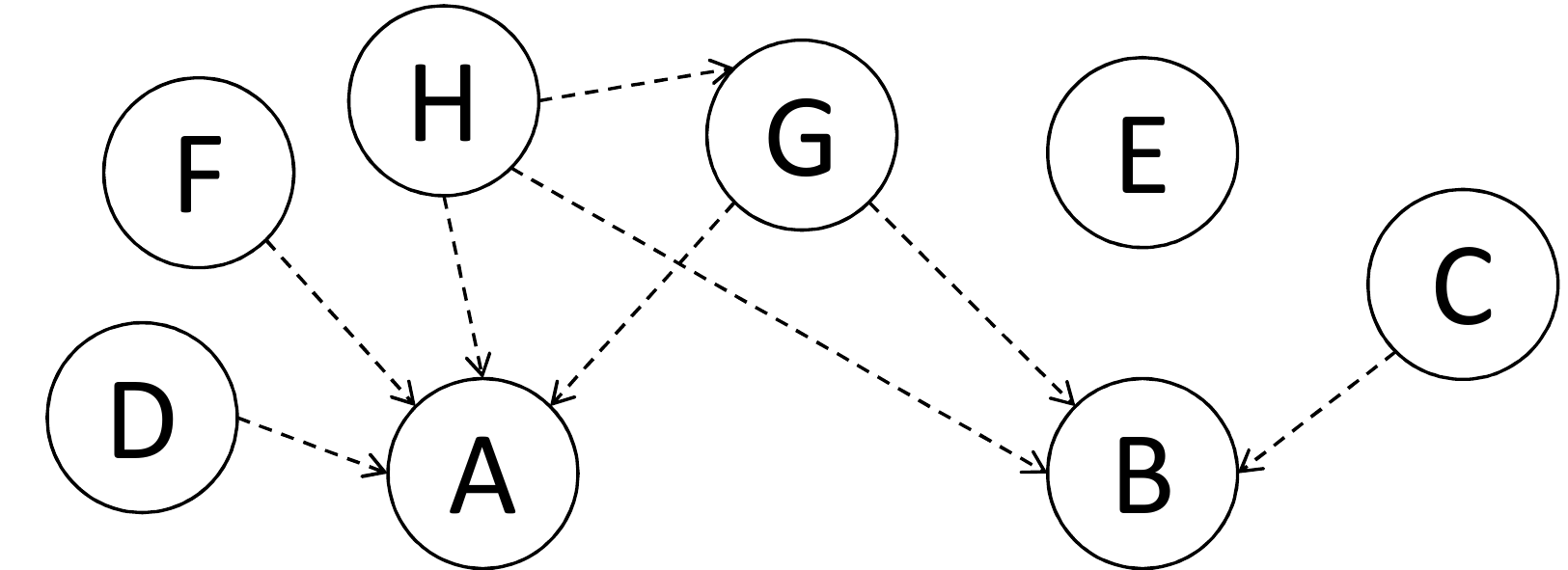} 
	\caption{Causality graph for Example \ref{ejm-causa}. Traced edges represent the causality relation.}
	\label{dep}
\end{figure}
\end{example}

\section{Lower and Upper Bounds of Extensions}
\label{LU-ext}

Section \ref{teoria} presented the definition of conflict-free ($\mathtt{cf}$) and admissible ($\mathtt{ad}$) sets and complete ($\mathtt{co}$), preferred ($\mathtt{pr}$), grounded ($\mathtt{gr}$), and stable ($\mathtt{st}$) semantics. Considering the causality graph, the arguments of an extension $\Ea_{\mathtt{x}}$ (for $\mathtt{x} \in \{\mathtt{cf, ad, co, pr, gr, st}\}$) may belong to $\mathtt{ARG}_\rightarrow$, $\mathtt{ARG}_\leftarrow$, or $\mathtt{ARG}_\circ$. Depending on it, the calculation of the probabilistic lower and upper bounds of each extension is different. Thus, we can distinguish the following cases: (i) the extension is empty, (ii) the extension has only one argument, and (iii) the extension includes more than one argument.

\begin{definition} \textbf{(Upper and Lower Bounds of Extensions)} Let $\Aa\Fa_{\Ca\Sa}=\langle \mathtt{ARG}, \Ra, \mathbb{K},f_{\Ca\Sa}\rangle$ be a Credal AAF,  $\mathbb{C} = \langle \mathtt{ARG}, \Ra_{\mathtt{CAU}} \rangle$ a causality graph, and $\Ea_{\mathtt{x}} \subseteq \mathtt{ARG}$ (for $\mathtt{x} \in \{\mathtt{cf, ad, co, pr, gr,st}\}$) an extension under semantics $\mathtt{x}$. The lower and uppers bounds of $\Ea_\mathtt{x}$ are obtained as follows:

1.  If $\Ea_\mathtt{x}=\{\}$, then $\underline{P}(\Ea_\mathtt{x})=0$ and $\overline{P}(\Ea_\mathtt{x})=1$, which denotes ignorance.\\
\indent2. If $|\Ea_\mathtt{x}|=1$, then $\underline{P}(\Ea_\mathtt{x})=\underline{P}(A)$ and $\overline{P}(\Ea_\mathtt{x})=\overline{P}(A)$ s.t. $A \in \Ea_\mathtt{x}$, where\\
\hspace*{0.2cm} $\underline{P}(A)$ and $\overline{P}(A)$ are obtained by applying Equation (1).\\
\indent3. If $|\Ea_\mathtt{x}|>1$, then $(\underline{P}(\Ea_\mathtt{x}),\overline{P}(\Ea_\mathtt{x}))= \mathtt{UL\_BOUNDS}(\Ea_\mathtt{x})$ (see Algorithm 1). 

Consider the following functions:

\indent- $f_\mathtt{CAU}(A)=\{B | (B,A) \in \Ra_\mathtt{CAU} \cup f_\mathtt{CAU}(B)\}$ \\
\indent- $\mathtt{TOP\_CAU}(\Ea_\mathtt{x})=\{A | A \in \mathtt{ARG}_{\leftarrow} \cap \Ea_\mathtt{x}\;\; and\;\;  \forall B\; s.t.\;  A \in f_\mathtt{CAU}(B), B \notin \Ea_\mathtt{x}\}$ \\
\indent- $\mathtt{FREE\_CAU}(\Ea_\mathtt{x})=\{A | A \in \mathtt{ARG}_{\rightarrow} \cap \Ea_\mathtt{x} \;\; and \;\; \forall B \in  f_\mathtt{CAU}(A), B \notin \Ea_\mathtt{x}\}$

\end{definition}

$\mathtt{TOP\_CAU}$ and $\mathtt{FREE\_CAU}$ consider only the arguments of $\Ea_\mathtt{x}$ and their causal relations restricted to $\Ea_\mathtt{x}$. The former returns the arguments that are caused by any of the other argument in $\Ea_\mathtt{x}$ but do not cause other argument(s) in $\Ea_\mathtt{x}$. If there is an argument that belongs to $\mathtt{ARG}_{\leftarrow}$ and $\mathtt{ARG}_{\rightarrow}$ in $\mathbb{C}$ but the argument(s) caused by it are not in $\Ea_\mathtt{x}$, then it is returned by $\mathtt{TOP\_CAU}$. The latter returns the arguments that belong to $\mathtt{ARG}_{\rightarrow}$ but whose caused arguments do not belong to extension $\Ea_\mathtt{x}$.

\begin{algorithm}
\label{algo-1}
\begin{algorithmic}[1]
\REQUIRE An extension $\Ea_\mathtt{x}$ and a causality graph $\mathbb{C}=\langle \mathtt{ARG}, \Ra_{\mathtt{CAU}} \rangle$
\ENSURE $(\underline{P}(\Ea_\mathtt{x}), \overline{P}(\Ea_\mathtt{x}))$
\IF {$(\Ea_{\mathtt{x}} \cap \mathtt{ARG}_{\leftarrow}) \neq \emptyset$}
	\STATE{$\mathtt{ARG}_{\leftarrow}''=\mathtt{TOP\_CAU}(\Ea_\mathtt{x})$ }
	\FOR{$i=1$ to $|\mathtt{ARG}_{\leftarrow}''|$}
		\STATE{$E_A^i=A \cup (f_{\mathtt{CAU}}(A) \cap \Ea_{\mathtt{x}})$ }
		\STATE{Calculate $K(E_A^i)$ \textit{//Calculate the credal set for} $E_A^i$\textit{ by 	applying Equation (3)}} 
	\ENDFOR
\ENDIF
\STATE{$\mathtt{ARG}_\circ' = \Ea_{\mathtt{x}} \cap \mathtt{ARG}_\circ$}
\IF{$(\Ea_{\mathtt{x}} \cap \mathtt{ARG}_{\rightarrow}) \neq \emptyset$}
	\STATE{$\mathtt{ARG}_\rightarrow'' = \mathtt{FREE\_CAU}(\Ea_\mathtt{x})$}
\ENDIF
\STATE{\textit{//* --- $\Ea_{\mathtt{x}}$ contains only one set of related arguments --- *//}}
\IF{$|\mathtt{ARG}_{\leftarrow}''|==1$ \&\& $\mathtt{ARG}_\circ'==\emptyset$ \&\& $\mathtt{ARG}_\rightarrow''==\emptyset$}
\STATE{\textit{// Apply Equation (4) for obtaining the lower and upper bounds of} $\Ea_{\mathtt{x}}$}
\STATE{$\underline{P}(\Ea_{\mathtt{x}})=\underline{P}(\mathtt{ARG}_{\leftarrow}'')$, $\overline{P}(\Ea_{\mathtt{x}})=\overline{P}(\mathtt{ARG}_{\leftarrow}'')$}
\ELSE
\STATE{\textit{//Apply Equation (2) for obtaining the lower and upper bounds of} $\Ea_{\mathtt{x}}$}
\STATE{$\underline{P}(\Ea_{\mathtt{x}})=\underline{P}(\bigcup_{i=1}^{i \leq |\mathtt{ARG}_{\leftarrow}''|} E_A^i \cup \mathtt{ARG}_\circ' \cup \mathtt{ARG}_\rightarrow'')$,}
\STATE{ $\overline{P}(\Ea_{\mathtt{x}})=\overline{P}(\bigcup_{i=1}^{i \leq |\mathtt{ARG}_{\leftarrow}''|} E_A^i \cup \mathtt{ARG}_\circ' \cup \mathtt{ARG}_\rightarrow'')$}
\ENDIF
\RETURN {$(\underline{P}(\Ea_{\mathtt{x}}, \overline{P}(\Ea_{\mathtt{x}})$}
\end{algorithmic}
\caption{Function $\mathtt{UL\_BOUNDS}$}
\end{algorithm}

\begin{example} (Cont. Example \ref{ejm-2} considering the causality graph of Example \ref{ejm-causa}). After applying the semantics presented in Definition \ref{def-semantics}, we obtain that $\Ea_{\mathtt{CO}}=\Ea_{\mathtt{PR}}=\Ea_{\mathtt{GR}}=\Ea_{\mathtt{ST}}=\{C,E,F,D,H,G\}$. Since this extension has more than one element, the Algorithm 1 has to be applied:

\vspace*{-0.3cm}

\begin{itemize}
\item We first evaluate the number of the caused arguments: $\Ea_\mathtt{y} \cap \mathtt{ARG}_\leftarrow = \{G\}$\break (for $\mathtt{y} \in \{\mathtt{CO, PR, GR, ST}\}$), then we obtain $\mathtt{TOP\_CAU}(\Ea_\mathtt{y})=\{G\}$ and $f_{\mathtt{CAU}}(G)= \{H\}$; hence, $E_G=\{G,H\}$. At last, we calculate the credal set for $E_G$ by applying Equation (3): $K(E_G)=\{0.7, 0.8, 1,0.9\}$. 

\item Next, we obtain those arguments that belong to the extension and that neither cause any other argument nor are caused by any other argument: $\mathtt{ARG}_\circ'=\{E\}$. 

\item Then, we evaluate the number of causing arguments: $\Ea_\mathtt{y} \cap \mathtt{ARG}_\leftarrow = \{C,D,F\}$ and we obtain $\mathtt{FREE\_CAU}(\Ea_\mathtt{y})=\{C,D,F\}$. 
\item Since $\Ea_\mathtt{y}$ do not contains only related arguments, we apply Equation (2) considering $K(E_G), K(E), K(C), K(D),$ and $K(F)$. 
\item Finally, we obtain: $(\underline{P}(\Ea_\mathtt{y}), \overline{P}(\Ea_\mathtt{y})= [0.0117,0.0806]$. 
\end{itemize}

Let us also take some conflict-free sets: $\Ea_{\mathtt{CF}}^1=\{A,F,H,D,E,G\}$,  $\Ea_{\mathtt{CF}}^2=\{A,F,H, D,G\}$, $\Ea_{\mathtt{CF}}^3=\{B,C,G,H\}$, and $\Ea_{\mathtt{CF}}^4=\{A\}$. The lower and upper bounds for these extensions are: $(\underline{P}(\Ea_\mathtt{CF}^1),\overline{P}(\Ea_\mathtt{CF}^1))=[0.13, 0.525]$, $(\underline{P}(\Ea_\mathtt{CF}^2),\overline{P}(\Ea_\mathtt{CF}^2))\break=[0.2,0.75]$, $(\underline{P}(\Ea_\mathtt{CF}^3),\overline{P}(\Ea_\mathtt{CF}^3))=[0.02,0.1875]$, and $(\underline{P}(\Ea_\mathtt{CF}^4),\overline{P}(\Ea_\mathtt{CF}^4))=[0.2,0.75]$.

\end{example}

So far, we have calculated the lower and upper bounds of extensions obtained under a given semantics. The next step is to compare these bounds in order to determine an ordering over the extensions, which can be used to choose an extension that resolves the problem. In this case, the problem was making a decision about a possible diagnosis between two alternatives: measles or chickenpox. We are not going to tackle the problem of comparing and ordering the extensions because it is out of the scope of this article; however, we can do a brief analysis taking into account the result of the previous example. Arguments $A$ and $B$ represent each of the alternatives. The unique extension under any semantics $\mathtt{y}$ does not include any of the alternatives. On other hand, free-conflict sets $\Ea_{\mathtt{CF}}^1$, $\Ea_{\mathtt{CF}}^2$ and $\Ea_{\mathtt{CF}}^4$ include argument $A$ and conflict free set $\Ea_{\mathtt{CF}}^3$ includes argument $B$. We can notice that there is a notorious difference between the lower and upper bounds of $\Ea_\mathtt{y}$ and the lower and upper bounds of any of the other conflict-free sets. In fact, the lower and upper bounds of the conflict-free sets have a better location. This may indicate that lower and upper bounds of extensions that include one of the alternatives are better than others of extensions that do not include any of the alternatives. This in turn indicates that using uncertainty in AAF may improve the resolutions of some problems, which was demonstrated in \cite{hunter2012some} for precise uncertainty and it is showed in the example by using imprecise uncertainty.

\section{Properties of the Approach}
\label{properties}

In this section, we study two properties of the proposed approach that guarantee (i) that the approach can be reduced to the AAF of Dung and (ii) that the values of both the lower and upper bounds of the extensions are between 0 and 1.

Given a Credal AAF $\Aa\Fa_{\Ca\Sa}=\langle \mathtt{ARG}, \Ra, \mathbb{K},f_{\Ca\Sa}\rangle$, $\Aa\Fa_{\Ca\Sa}$ is \textit{maximal} if $\forall A \in \mathtt{ARG}$ it holds that $p_i =1$ ($1 \leq i \leq m$) where $p_i \in K(A)$ and $K(A)=f_{\Ca\Sa}(A)$ and $\Aa\Fa_{\Ca\Sa}$ is \textit{uniform} if $0 \leq p_i \leq 1$. Be maximal transforms an $\Aa\Fa_{\Ca\Sa}$ into a standard AAF of Dung, which means that every agent believes that every argument is believed without doubts. The next proposition shows that a $\Aa\Fa_{\Ca\Sa}$ can be reduced to an AAF that follows Dung's definitions. 

\begin{proposition} Given a credal AAF $\Aa\Fa_{\Ca\Sa}=\langle \mathtt{ARG}, \Ra, \mathbb{K},f_{\Ca\Sa}\rangle$ and a extension $\Ea_\mathtt{x}$ $(\mathtt{x} \in \{\mathtt{cf, ad, co, pr, gr, st}\})$. If $\Aa\Fa_{\Ca\Sa}$ is maximal, then $\forall \Ea_\mathtt{x} \subseteq \mathtt{ARG}$, $\underline{P}(\Ea_\mathtt{x})=\overline{P}(\Ea_\mathtt{x})=1$.

\end{proposition}

\begin{proof} Since $\Aa\Fa_{\Ca\Sa}$ is maximal, then $\forall A \in \mathtt{ARG}, K(A)=\{1_1,...,1_m\}$.
In order to obtain the $\underline{P}(\Ea_\mathtt{x})$ and $\overline{P}(\Ea_\mathtt{x})$, Equations (1), (2), or (4) have to be applied. For Equation (1): the $inf\{1,...,1\}=sup \{1,...,1\}=1$. For Equation (2): $\forall A, \prod \{1,...,1\}=1$, so the minimum and maximum of a set composed of 1s is always 1. The same happens with Equation (4). 
\end{proof}

\begin{proposition} Given a credal AAF $\Aa\Fa_{\Ca\Sa}=\langle \mathtt{ARG}, \Ra, \mathbb{K},f_{\Ca\Sa}\rangle$ and a extension $\Ea_\mathtt{x}$ $(\mathtt{x} \in \{\mathtt{cf, ad, co, pr, gr, st}\})$. If $\Aa\Fa_{\Ca\Sa}$ is uniform, then $\forall \Ea_{\mathtt{x}} \subseteq \mathtt{ARG}$, $0 \leq \underline{P}(\Ea_{\mathtt{x}}) \leq 1$ and $0 \leq \overline{P}(\Ea_{\mathtt{x}}) \leq 1$.
\end{proposition}

\begin{proof} In order to obtain the $\underline{P}(\Ea_\mathtt{x})$ and $\overline{P}(\Ea_\mathtt{x})$, Equations (1), (2), or (4) have to be applied. Since $\Aa\Fa_{\Ca\Sa}$ is uniform, we can say that the minimums (infimums) and maximums (supremums) are always between 0 and 1. Besides, the product of two numbers between 0 and 1 is always between 0 and 1.

\end{proof}

\section{Related work}
\label{relacionado}
In this section, we present the most relevant works -- to the best of our knowledge -- that study probability and abstract argumentation. These works assign probability to the arguments, to the attacks, or to the extensions and all of them use precise probabilistic approaches. Thus, as far as we know, we are introducing the first abstract argumentation approach that employs imprecise probabilistic approaches.

Dung and Thang \cite{dung2010towards} propose an AF for jury-based dispute resolution, which is based on probabilistic spaces, from which are assigned probable weights -- between zero and one -- to arguments. In the same way, Li et al. \cite{li2011probabilistic} present and extension of Dung's original AF by assigning probabilities to both arguments and defeats. Hunter \cite{hunter2012some} bases on the two articles previously presented and focuses on studying the notion of probability independence in the argumentation context. The author also propose a set of postulates for the probability function regarding admissible sets and extensions like grounded and preferred. Following the idea of using probabilistic graphs, the author assigns a probability value to attacks in \cite{hunter2014probabilistic}.

Thimm \cite{thimm2012probabilistic} focuses on studying probability and argumentation semantics. Thus, he proposes a probability semantics such that instead of extensions or labellings, probability functions are used to assign degrees of belief to arguments. An extension of this work was published in \cite{thimm2017probabilities}. Gabbay and Rodrigues \cite{gabbay2015probabilistic} also focus on studying the extensions obtained from an argumentation framework. Thus, they introduce a probabilistic semantics based on the equational approach to argumentation networks proposed in \cite{gabbay2012equational}.

\section{Conclusions and Future Work}
\label{conclus}

This work presents an approach for abstract argumentation under imprecise probability. We defined a credal AAF, in which credal sets are used to model the the uncertainty values of the arguments, which correspond to opinions of a set of agents about their degree of believe about each argument. We have considered that -- besides the attack relation -- there also exists a causality relation between the arguments of a credal AAF. Based on the credal sets and the causality relation, the lower and upper bounds of the extensions -- obtained from a semantics -- are calculated. 

We have done a brief analysis about the problem of comparing and ordering the extensions based on their lower and upper bounds; however, a more complete analysis and study are necessary. In this sense, we plan to follow this direction in our future work. We also plan to further study the causality relations, more specifically in the context of credal networks \cite{cozman2000credal}. Finally, we want to study the relation of this approach with bipolar argumentation frameworks \cite{amgoud2008bipolarity}.

\section*{Acknowledgment}
This work is partially founded by CAPES (Coordena\c{c}\~{a}o de Aperfei\c{c}oamento de Pessoal de N\'{i}vel Superior).

\bibliographystyle{unsrt}  
\bibliography{mybibliography2}  


\end{document}